# A Neurorobotics Approach to Investigating the Emergence of Communication in Robots

Jungsik Hwang
Cognitive Neurorobotics Research Unit
Okinawa Institute of Science and Technology
Okinawa, Japan
jungsik.hwang@gmail.com

Nadine Wirkuttis
Cognitive Neurorobotics Research Unit
Okinawa Institute of Science and Technology
Okinawa, Japan
nadine.wirkuttis@oist.jp

Jun Tani
Cognitive Neurorobotics Research Unit
Okinawa Institute of Science and Technology
Okinawa, Japan
tani1216jp@gmail.com

*Abstract*—This paper introduces our approach to building a robot with communication capability based on the two key features: stochastic neural dynamics and prediction error minimization (PEM). A preliminary experiment with humanoid robots showed that the robot was able to imitate other's action by means of those key features. In addition, we found that some sorts of communicative patterns emerged between two robots in which the robots inferred the intention of another agent behind the sensory observation.

## I. INTRODUCTION

The ability to communicate with others is one of the most fundamental skills for many animals, including humans. In this paper, we introduce our approach to building a socially interactive agent based on two key principles: stochastic neural dynamics and prediction error minimization (PEM). Dealing with fluctuating signal is essential for the robot to learn from noisy sensorimotor experience. PE minimization is at the core of the theory of predictive coding and it has been shown that intention behind sensory observation can be inferred by minimizing prediction error (e.g., the discrepancy between sensory observation and prediction) [1]. Based on these principles, we implemented a dynamic neural network model and conducted a set of synthetic robotic experiments to examine the emergence of communication between the agents.

A few recent studies have illustrated similar approaches. In [2], they found that the PE minimization mechanism elicited the adaptive behavior of the robot and afforded the robot to recover from deformed interaction caused by external instability. In [3], the same robotic platform was used and the neural network model with precision-weighted PE minimization mechanism was examined. They found that the inflexible behaviors of the robot were caused by aberrant sensory precision.

Our approach differs from previous works in three important ways. First, we used the neural network model [4] which can develop stochastic dynamics to deal with fluctuating sensory signals. Second, the other's behavior is explicitly represented in our experiment. The previous studies [2, 3], however, used the XY coordinate of the ball as the input to represent other agent's action, undermining communication. For instance, "attract" gesture in their studies were not, indeed, observed by another agent, meaning that the robots were not communicating with each other. Third, we examined the effect of the model parameters from the training process whereas the previous study [3] modulated the model's parameter (sensory precision) only after training. Consequently, their approach could not illustrate the influences of sensory precision on learning (neuroplasticity).

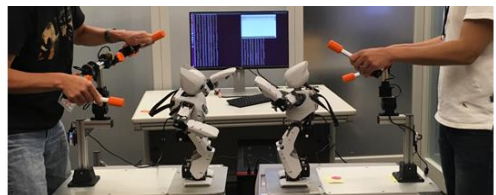

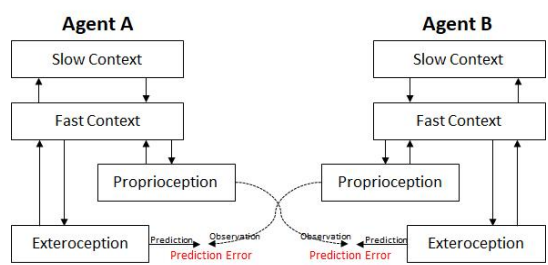

Fig. 1. Experiment setting. (a) an example of human-human interaction and (b) a diagram illustrating interaction between the two neural network models.

## II. NEURAL NETWORK MODEL

A stochastic hierarchical neural network model based on [4] was implemented (Fig. 1). A previous study [4] showed that the model can develop stochastic dynamics which improves the learning performance, especially with fluctuating data. The model consists of a set of context layers showing different (fast and slow) timescale neural dynamics (Fig. 1 (b)). The model predicts two different types of signal: proprioceptive and exteroceptive signals indicating the robot's own and other's action respectively. During training, the model is trained to generate predictions of both signals and the learnable parameters are optimized in the direction of minimizing prediction error.

The neural network model used in this study has two important features. First, the level of stochasticity of neural activation can be controlled by a parameter W. In brief, higher W forces the neural activation to follow the unit Gaussian distribution, meaning that the past information encoded in the internal states would be removed and the neural activation would be more stochastic. Second, the model provides a PEM mechanism. In this scheme, the neural dynamics is updated to minimize PE at the output layer. Previous studies [1, 5] showed that the underlying intention behind the observed patterns can be inferred by minimizing PE.

## III. EXPERIMENT AND RESULTS

Two small humanoid robots (ROBOTIS OP2) were used in our experiments. Prior to training, the communication patterns between the human operators were collected. Then, the model was trained with the collected data and it was tested in different situations including Human-Robot and Robot-Robot Interaction.

### A. Data Collection from Human-Human Interaction (HHI)

Prior to training, two operators interacted each other through the robots (Fig. 1 (a)). In order to control the robot, we used the controller which shared the same joint configuration with OP2. During HHI, the encoder values of both controllers were collected to represent proprioceptive signal (i.e. own action) and exteroceptive signal (i.e. other's action). A total number of 62 sequences consisting of 31 gesture imitation tasks under 2 different roles (leading and following the imitation) were collected.

### B. Imitation Capability in Human-Robot Interaction (HRI)

In this setting, we assumed a typical HRI setting where the communication between the user and the robot can be examined. To compare the model's performance systematically, we provided the same pre-recorded gestures (short and long data) to the models with different parameters and compared the imitation performance measured as the correlation coefficient between other's gesture and own gesture.

The result (Table I) revealed that the model with stochastic dynamics performed better than the one with deterministic dynamics. Particularly, more deterministic higher level condition (ID 5) performed the best in both short and long gesture cases. In contrast, the deterministic condition (ID 1) showed the worst performance. This result highlights the importance of embedding stochasticity as suggested in [4]. In addition, we also found that the model generally performed better in the PEM condition, indicating the importance of PEM to adapt to fluctuating signals.

### C. Emergence of Communication in Robot-Robot Interaction (RRI)

In this setting, each robot is operated by a neural network model and the robots interact with each other. In each trial, the models interacted for 1,000 steps and we conducted 10 trials. Then, we examined what sorts of communication patterns emerged between two agents.

Fig. 2 illustrates the different types of communication emerged between the two robots. We found that PEM generally initiated some sorts of social interaction. The effect of W was less obvious than the effect of the PEM. When both agents were not performing PEM, the communication patterns mostly converged to either fixed point (i.e. no social interaction) or limit cycle (i.e. repetitive behavior). When at least one of the agents was performing PEM, emergent interactions were observed. Although "meaningful" social interaction is not strictly defined in this context, it can be still observed that inferring other's intention through minimizing PE elicited social interaction whereas the model without PEM (i.e. passive perception without inference) showed less social behavior, such as repetitive or no actions at all.

TABLE I. THE IMITATION PERFORMANCE OF THE MODEL

| | | Network Conditions | | | | |
|---|---|---|---|---|---|---|
| | W | (H) 0.0 (L) 0.0 | (H) 0.0001 (L) 0.0001 | (H) 0.01 (L) 0.01 | (H) 0.01 (L) 0.0001 | (H) 0.0001 (L) 0.01 |
| | ID | 1 | 2 | 3 | 4 | 5 |
| Short Data | NO PEM | 0.48 | 0.65 | 0.67 | 0.66 | 0.68 |
| | PEM | 0.61 | 0.66 | 0.68 | 0.67 | 0.69 |
| Long Data | NO PEM | 0.55 | 0.68 | 0.69 | 0.68 | 0.69 |
| | PEM | 0.65 | 0.68 | 0.69 | 0.69 | 0.70 |

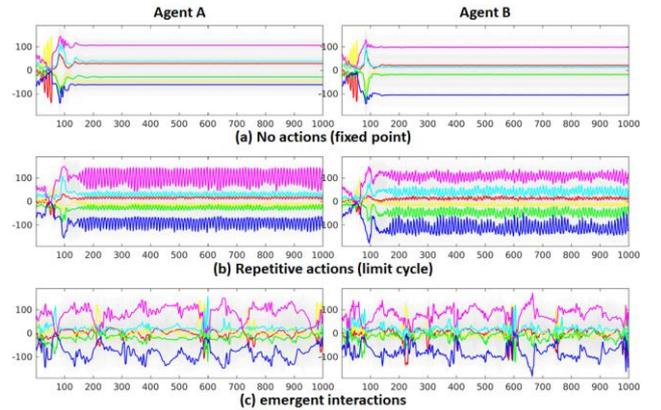

Fig. 2. Three examples of communication patterns between two agents. X and Y axis represent the times steps and joint position values respectively. The colors represent 6 DoFs of the robot used in our study. (a) no actions and (b) repetitive actions between the agents and (c) emergent communicative patterns between the agents

## IV. CONCLUSION

In this paper, we introduced our approach to building an artificial agent with communicative skills based on the two key principles: stochastic neural dynamics and prediction error minimization (PEM). The preliminary experiment results showed that the neural network model with stochastic dynamics was able to deal with fluctuating communication patterns. In addition, we found that the ability to infer other's intention by minimizing PE played an essential role in communication with another agent. The model imitated the gestures better with PEM. Furthermore, it induced the emergence of communication between two artificial agents.